\documentclass[letterpaper, 10 pt, conference]{ieeeconf}  

\IEEEoverridecommandlockouts                              

\usepackage{mathptmx} 
\usepackage{times} 
\usepackage{amsmath} 
\usepackage{amssymb}  
\usepackage{tabularx} 
\usepackage{graphicx}
\usepackage{comment}
\usepackage{mathtools}
\usepackage{upgreek}
\usepackage{gensymb}
\usepackage[noadjust]{cite}
\usepackage{placeins}
\usepackage{xcolor}
\usepackage{url}
\usepackage{mwe}
\usepackage{balance}
\usepackage{multirow}

\DeclarePairedDelimiter\abs{\lvert}{\rvert}%
\DeclarePairedDelimiter\norm{\lVert}{\rVert}%

\makeatletter
\let\oldabs\abs
\def\abs{\@ifstar{\oldabs}{\oldabs*}}
\let\oldnorm\norm
\def\norm{\@ifstar{\oldnorm}{\oldnorm*}}
\makeatother

\title{\LARGE \bf
Characterization and Evaluation of Screw-Based Locomotion Across Aquatic, Granular, and Transitional Media
}

\author{Derek Chen$^1$, Zoe Samuels$^2$, Lizzie Peiros$^1$, Sujaan Mukherjee$^2$, Michael C. Yip$^1$ \IEEEmembership{Senior Member, IEEE}
\thanks{$^1$ Derek, Lizzie, and M.C. Yip are with the Electrical and Computer Engineering Department, University of California San Diego, La Jolla, CA 92093 USA. {\tt\footnotesize\{dec004, epeiros, yip\}@ucsd.edu}}%
\thanks{$^2$ Zoe and Sujaan, are with the Mechanical and Aerospace Engineering Department, University of California San Diego, La Jolla, CA 92093 USA. {\tt\footnotesize \{zsamuels, s2mukherjee\}@ucsd.edu}}%
}

\begin{document}

\maketitle
\thispagestyle{empty}
\pagestyle{empty}

\begin{abstract}
Screw-based propulsion systems offer promising capabilities for amphibious mobility, yet face significant challenges in optimizing locomotion across water, granular materials, and transitional environments. This study presents a systematic investigation into the locomotion performance of various screw configurations in media such as dry sand, wet sand, saturated sand, and water. Through a principles-first approach to analyze screw performance, it was found that certain parameters are dominant in their impact on performance. Depending on the media, derived parameters inspired from optimizing heat sink design help categorize performance within the dominant design parameters. Our results provide specific insights into screw shell design and adaptive locomotion strategies to enhance the performance of screw-based propulsion systems for versatile amphibious applications.
\end{abstract}

\section{Introduction}

In spaces inaccessible to humans or that have safety concerns for human intervention, robots are often used in place of a physical person for ecological surveys, as well as for data collection. This challenge is particularly evident in applications requiring access to confined or hazardous spaces. For example, geologists in Hawaii wishing to study lava tubes cannot safely send in human divers as the environment is too narrow for human exploration and contains many hazards \cite{sawlowicz2020short}. The lava tube environment is multi-modal, necessitating mobility across water, sand, and transition areas between those two terrains \cite{rotz2023}. While considerable advancements have been made in robotic mobility within singular domains, such as terrestrial or aquatic locomotion, the development of systems capable of seamlessly navigating diverse environments remains an active and critical area of research \cite{rafeeq2021}.

Multiple robotic platforms have been invented to traverse various unstructured environments, some of which have multi-terrain capabilities. The required environments for lava caves and marshlands include water, sand, plant matter, and granular media. Several robotic systems have demonstrated capabilities in these individual materials \cite{kashem2019, baines2022multi, zhang2013initial}, though practical limitations often emerge in real-world applications. While laboratory studies have shown that certain designs can operate in these materials, the velocity achieved is often low, and environmental challenges limit the operational permanence due to design constraints. In practice, these theoretical capabilities frequently do not translate to reliable field performance.

\begin{figure}
    \centering
    \includegraphics[width=\linewidth]{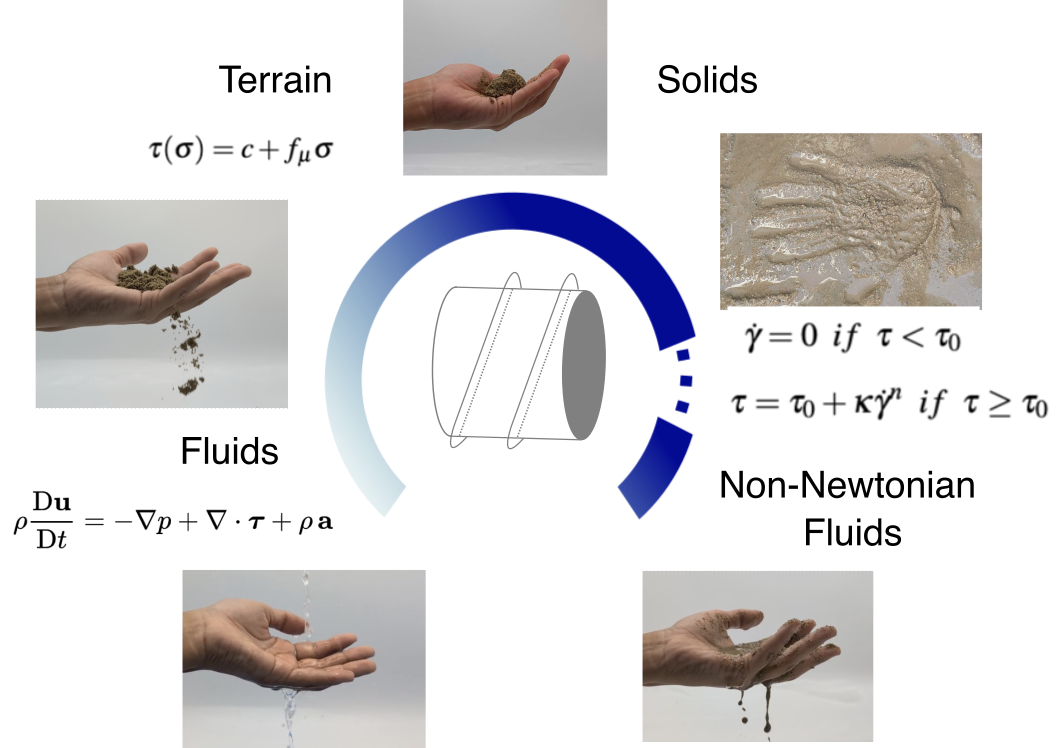}
    \caption{The figure above shows the full spectrum of materials and their models. Starting from the lower left, fluids use classical models, including the Navier-Stokes equations, which describe mass transport of material through a control volume. Next in the upper left is the Terrain model that describes how cohesion and normal force impact shear. Next are solids, which include one of the operating modes of saturated sand, and finally, in the bottom right are non-Newtonian fluids that include a second operating mode where shear changes due to agitation.}
    \label{fig:spectrum}
\end{figure}

Archimedes' screw locomotion currently shows promise for achieving locomotion in both terrestrial and aquatic environments as field robots, with the form factor and space-saving constraints allowing mobility within environmental challenges \cite{arcsnake_icra, Ko2020}. Studies have demonstrated the promising performance of Archimedes screws with screw propulsion in granular and aquatic media, with multiple screw-driven robots taking advantage of this foundational knowledge. Robotic platforms such as ARCSnake have demonstrated the practical application of screw propulsion for multi-domain mobility, including effective aquatic locomotion \cite{arcsnake_icra}.

Despite the effectiveness of Archimedes screw robots in aquatic and granular media, they have struggled when transitioning between certain media despite their relatively strong performance in each individual medium. This limitation stems from a lack of understanding of the transition zones and how different parameters interact between different media. Many studies that include water and sand do not address transitional media such as wet sand or saturated sand conditions, creating gaps in the understanding of multi-modal terrain navigation. Empirical results from such platforms, such as slow speed in different media, demonstrate clear limitations in performance across multi-modal terrain, highlighting the necessity for better performing screws to enable enhanced functionality \cite{arcsnake_icra}. The Archimedes screw faces a critical limitation in performance optimization, as varying certain design parameters like screw depth produces different effects on efficiency in aquatic media versus granular media, making unified design approaches challenging.

Previous works focus on screw parameters from engineering design, such as angle of attack (pitch) to optimize performance, and found that relationships to locomotive performance are not clearly defined. Our work presents a comprehensive experimental study focusing on screw-based locomotion across aquatic, transitional media, and land-based environments. By systematically analyzing performance metrics under varying conditions, we demonstrate that certain design parameters form the dominating metric of performance in certain media. In places where the dominating design parameters match, derived parameters like aspect ratio, crowdedness, tip speed (or mass transport), and sinkage, which are formed as functions of engineering parameters like pitch, blade height, and number of starts, provide better context to analyze locomotive performance than traditional engineering parameters alone. Furthermore, we provide alternative modes of locomotion (rolling) in transitional media as traditional screw models break down. This approach provides better metrics for the design of screw shells in amphibious Archimedes screw-driven robots, offering a better framework for multi-terrain robotic locomotion systems and provide beneficial strategies for design optimization.


\section{Related Works}

Recent literature reviews have established the systematic development of Archimedean screw propulsion from early amphibious traction in farmland to contemporary robotic implementations across more varied marshy terrain \cite{villacres2023literature}. This historical perspective highlights persistent challenges in optimizing performance across diverse media.

Initial theoretical models made to help characterize different screw propulsion in deformable media, including water, made simplifying assumptions such as neglecting energy losses due to media displacement and assuming no-slip conditions \cite{Cole_1961, dugoff_1967}. Subsequent research in terra-mechanics has provided more sophisticated mathematical frameworks for representing screw modeling to account for slip and energy dissipation during media interaction \cite{ishigami2009slip, Nagaoka_2010}. The introduction of such terra-mechanics-based propulsive characteristics helped to address soil-vehicle interactive mechanics. These models extend beyond simple assumptions to incorporate complex interactions between screw geometry and surrounding media properties. However, the terra-mechanics models were primarily developed for large-scale vehicles operating in specific uniform soil conditions \cite{he2019review}, limiting their application for smaller-scale robotic systems, as they often fail to accurately predict dynamic behavior or capture the interactions in non-homogenous and transitional media.

Recent research has begun to address this gap in aquatic propulsion analysis. For instance, \cite{engproc2023033032} performed computational simulations to analyze the hydrodynamic interactions of fully submerged Archimedean screws, examining how variations in helix angle and advance ratio affect thrust and torque distribution. Their findings highlight the influence of screw geometry on propulsion efficiency in water.

Additionally, studies have examined screw-generated forces in granular media through experimental, computational, and analytical comparisons \cite{marvi_2019, lan2024dem}. Other studies have looked into empirical testing to extrapolate design parameters to tweak and adjust for optimal performance. Seo et al. \cite{seo2021robust} delved into the extensive testing of different screw parameters, such as blade height and profile in granular media. Past experiments by Lim et al. \cite{lim2023mobilityanalysisscrewbasedlocomotion, jason_thesis} have investigated the effects of Archimedean screw propulsion in granular and later aquatic situations, but have not investigated whether the resulting changes in efficiency were better represented by derived parameters such as crowdedness or relative velocity.

Controlled tests in existing literature don't consider the transition region between media, particularly the complex transition from sand to water, specifically addressing how water affects the properties of sand and locomotion strategies. Previous works focused on a single screw parameter from engineering design, angle of attack (pitch), and could not conclude a relationships to locomotive performance across all media \cite{lim2023mobilityanalysisscrewbasedlocomotion}.

Our work addresses these limitations in transitional media and multi-domain movement by presenting a comprehensive experimental study focusing on screw-based locomotion in aquatic, transitional media, and land-based environments. 

\section{Methods}

In this paper, we present a new framing for selecting screw parameters that influence screw propelled robot strategy across multimedia, which is not accurately captured with the singular granular media/terra-mechanics model. We propose looking at multi-domains, looking at a spectrum of models that better capture the observable behavior of materials. This spectrum is depicted in Figure \ref{fig:spectrum}. First, we present a new framing for modeling the screw interactions in fluids that affect Archimedean screw's performance by drawing parallels to optimizing heat sinks (Figure \ref{fig:heat_sink}). Second, we show a more nuanced means of applying terra-mechanics models to material that is modeled as both a fluid and terrain. Finally, we present a non-Newtonian fluid that does not follow any traditional modeling involving screws.

In this section, four media are described: water, dry sand, wet sand, and saturated sand. 
Each of the four categories has a different model or combination of models to predict ideal locomotion performance. 
Additionally we present a transitional set of parameters that connect the traditional screw design parameters, blade height, pitch, and number of starts, to the modeling equations that govern performance. 
These parameters, blade spacing, tip speed, correlated to mass transport, and sinkage, will be referred to as derived parameters, as they are derived from the mathematics and are a combination of the screw design parameters. 

\subsection{Thrust as a Function of Screw Parameters}

Numerous parameters have varying effects on screw performance; however, the trends are not clear, nor do they hold with adjustments to other parameters. The parameters in question are shown in a mathematical proportion to thrust, as shown below:
\begin{equation}
    F_{max} \underset{\sim}{\propto} f(B_H, \alpha, N, m, \omega_s, c, f_{\mu})
\end{equation}
where $F_{max}$ is the maximum thrust force, $B_H$ is the blade height, $\alpha$ is the pitch angle, $N$ is the number of starts, $m$ is the mass of the screw, $\omega_s$ is the angular velocity of the screw, $c$ is the cohesion of the material being moved through, and $f_{\mu}$ is the friction angle coefficient of the material being moved through. 
Of these variables, only some are directly controlled in the design, $B_H$, $\alpha$, $N$, $m$, $\omega_s$, and are shown in Fig. \ref{fig: Derived_Design_Param}.

Using classical models from multiple domains as a guide, more complex derived parameters are proposed to show the correlation with certain groupings of parameters and performance. 

\subsection{Classical Models}
\begin{figure}[t!]
    \vspace{2mm}
    \centering
    \includegraphics[width=\linewidth]{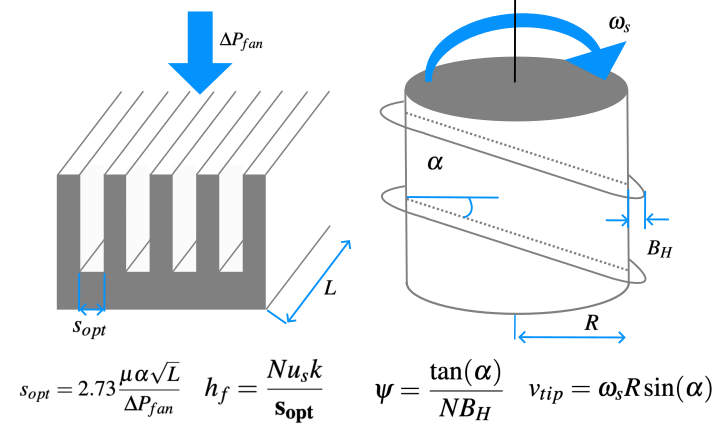}
    \caption{In Figure 2 are labeled drawings of a generic heat sink and a generic screw. The equations for optimizing heat transfer include the equation that defines the optimal spacing of the fins. These equations inspired a new way of thinking about screw parameters to create an improved intuition about performance using aspect ratio (geometry) and tip speed (mass transport).}
    \label{fig:heat_sink}
\end{figure}

\subsubsection{Heat Sink Modeling}

In heat sink design there are a few fundamental parameters which are controllable: geometry and velocity of air controlled through a pressure gradient. The forced air model designates an ideal fin spacing based on these parameters \cite{bejan1992optimal,teertstra2000analytical,kays1984compact,muzychka1998modeling} as shown in Figure \ref{fig:heat_sink} where $s_{opt}$ is the spacing of the heat sink fins, $L$ is the length of the heat sink normal to the fin spacing distance, $\Delta P_{fan}$ is the pressure gradient across the heat sink which derives the flow rate or speed of air flowing through the spaced channels and overall geometry, and $\mu\alpha$ is the property group which is material specific. This optimal spacing then plays a part in characterizing the average heat transfer co-efficient ($h_f$), which is maximized for a heat sink \cite{bejan1992optimal,teertstra2000analytical,kays1984compact,muzychka1998modeling} as shown in Figure \ref{fig:heat_sink}.

\begin{figure*}[t!]
    \vspace{2mm}
    \centering
    \includegraphics[width=0.95\linewidth]{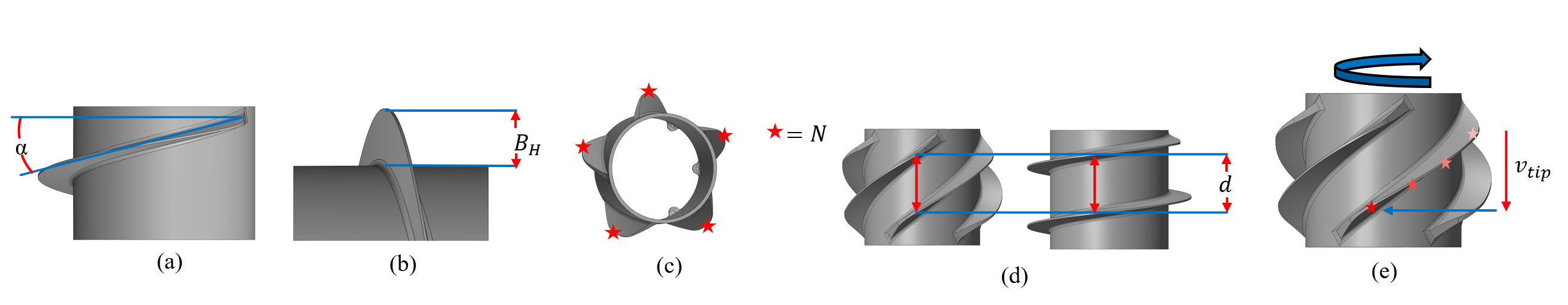}
    \caption{(a) Pitch angle ($\alpha$) is taken to the be the angle of attack of the screw (b) Blade height ($B_H$) is the distance between the inner circular housing and the tip of the blade (c) Number of starts ($N$) is the number of unique blades as seen from the front of the screw shell (d) Blade spacing ($d$) is the spacing between blades seen from a side view (e) Tip speed ($v_tip$) is the speed a fixed reference point moves on the blade when the screw shell is traveling}
    \label{fig: Derived_Design_Param}
\end{figure*}

\subsubsection{Terra-mechanics}
Terra-mechanics models are drawn from the mechanics of materials: stress, strain, shear, and friction. Screw locomotion is dominated by shear stress. The most simplified equation is shown below:

\begin{equation}
\label{beam}
    \tau = \frac{V}{A}
\end{equation}
where $\tau$ is shear stress, $V$ is shear force, and $A$ is surface area. For most applications, this equation is sufficient, however, when moving through granular media, more complex models better capture the behavior, as shown in Figure \ref{fig:spectrum} and demonstrated in \cite{huang2023influence}.
Where $\sigma$ is the normal stress, $c$ is the cohesion of the material, and $f_{\mu}$ is the friction angle coefficient. This equation shows the relationship between normal loading and material properties like cohesion, the tendency of particles to stick together (influenced by moisture), as they impact the maximum shear of the material.

In the mechanics of screw propulsion. Shear plays a fundamental role in two areas: (1) the rigidity of the grooves for the screw blades to propel through (2) the amount of torque required to cut into fresh material to create the grooves. 


\subsubsection{Non-Newtonian Fluid}

Saturated sand is categorized as a shear-thinning fluid or pseudo-plastic fluid with non-Newtonian viscosity. This means that the apparent viscosity decreases with increased stress. When unagitated, the saturated sand behaves like a solid; however, tides in the ocean, which bring water to the surface, decrease the shear, and the solid beach sand becomes quicksand. This phenomenon is described in the Herschel-Bulkley model \cite{chambon2014experimental, bates2017dam} and shown in Figure \ref{fig:spectrum} where $\tau$ is the shear stress, $\tau_0$ is the yield stress, $\kappa$ is the consistency index, $\dot\gamma$ is the shear rate, and $n$ is the flow index, which for shear thinning is equal to 1. A $\tau_0 = 1$ would be a Newtonian fluid.


\subsection{Relationship of derived and design parameters}

In screw design, the parameters include: blade height, pitch, and number of starts depicted in Figure \ref{fig: Derived_Design_Param} (a-c). Models as this time have a difficult time telling a full story of how to predict performance, particularly across media. Given the complexity and variability of models for relevant media, this new proposed perspective of derived parameters better explains the overall performance of stress across variable materials.

The derived parameters explored in this work include: aspect ratio and tip speed.

\subsubsection{Aspect Ratio}
Going off the basis that there is an optimal spacing $s_{opt}$ from heat sink modeling, the Aspect Ratio is a dimensionless number that helps correlate the number of starts, N, pitch angle, $\alpha$, and blade height $B_h$. A higher value corresponds to a screw that has a higher blade spacing relative to blade height, while a lower value corresponds to a more crowded design, with lower blade distance and higher blade height. 
\begin{equation}
    \psi = \frac{\tan(\alpha)}{N B_H}
\end{equation}

\subsubsection{Theoretical Tip Speed}
Tip speed $v_{tip}$ is the relative speed of material pushed backwards and is calculated as:

\begin{equation}
    v_{tip} = \omega_s R \sin(\alpha)
\end{equation}

Where $\omega_s$ is the angular velocity of the screw, $alpha$ is the pitch angle, and $R$ is the radius of the screw body.

For media that are granular, such as sand or wet sand, the predominant factor contributing to $F_{max}$ is expected to be blade height $B_H$. With higher blade height comes higher normal stress $\sigma$ in the sand due to the increased weight of sand compacting the lower layers of sand, leading to higher shear stress and better grooves for the blade to push off of.

In liquids, pitch angle $\alpha$ becomes the driving factor in $F_{max}$, with the higher velocity contributing to the equivalent of a higher pressure gradient $\Delta P_{fan}$, thus increasing the performance of the screw shell.

For the non-Newtonian saturated sand, $F_{max}$ provides no relationship as the screw fails to turn in the media, with the sand being treated as solid on impact. In such a scenario, an alternative method of locomotion, such as rolling on the surface, proves more effective.

\section{Experiments and Results}

\begin{table}[t]
\centering
\caption{Screw Shell Testing Parameters}
\begin{tabular}{|l|c|l|c|l|}
\hline
\textbf{Shell} & \textbf{Blade Height} & \textbf{Pitch Angle} & \textbf{\# Starts} & \textbf {Aspect Ratio}\\
\hline
111 & Low (10mm) & Low (8.4$^{\circ}$) & 1 & 0.015\\
\hline
311 & High (30mm) & Low (8.4$^{\circ}$) & 1 & 0.005\\
\hline
122 & Low (10mm) & Med (20.1$^{\circ}$) & 2 & 0.018\\
\hline
322 & High (30mm) & Med (20.1$^{\circ}$) & 2 & 0.006\\
\hline
135 & Low (10mm) & High (36.2$^{\circ}$) & 5 & 0.015\\
\hline
335 & High (30mm) & High (36.2$^{\circ}$) & 5 & 0.005 \\
\hline
Variable & High (30) & (2.7$^{\circ}$-19.2$^{\circ}$ ) & 2 & ( - )\\
\hline
\end{tabular}
\label{table:Testing Parameters}
\end{table}

\begin{figure*}[t]
    \vspace{2mm}
    \centering
    \includegraphics[width=0.95\linewidth]{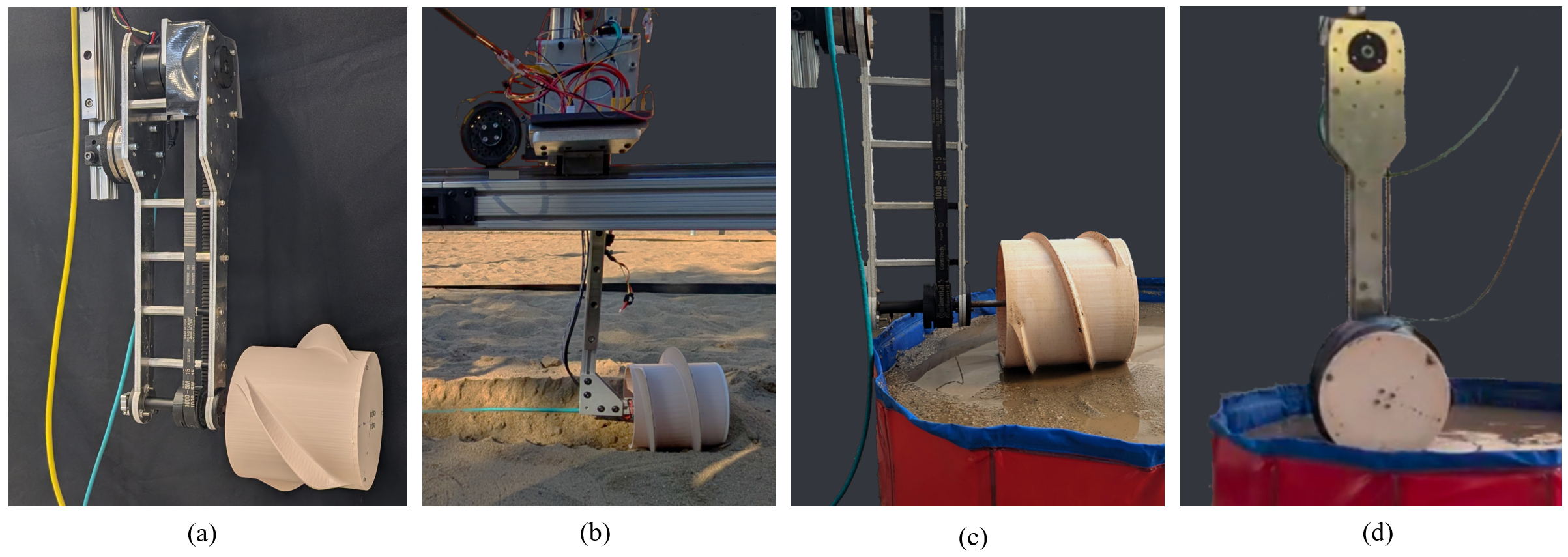}
    \caption{(a) The configuration of the testbed for aquatic testing, using the aquatic adapter to keep electronics out of the water (b) The configuration for sand and wet sand testing, where the motor is integrated inline with the screw shell to give better stiffness metrics. (c) Saturated sand testing uses an identical setup to sand testing. (d) Rolling set up where linear movement of the screw is generated through rolling.}
    \label{fig:testsetup_temp}
\end{figure*}

\begin{figure}
    \centering
    \includegraphics[width=0.85\linewidth]{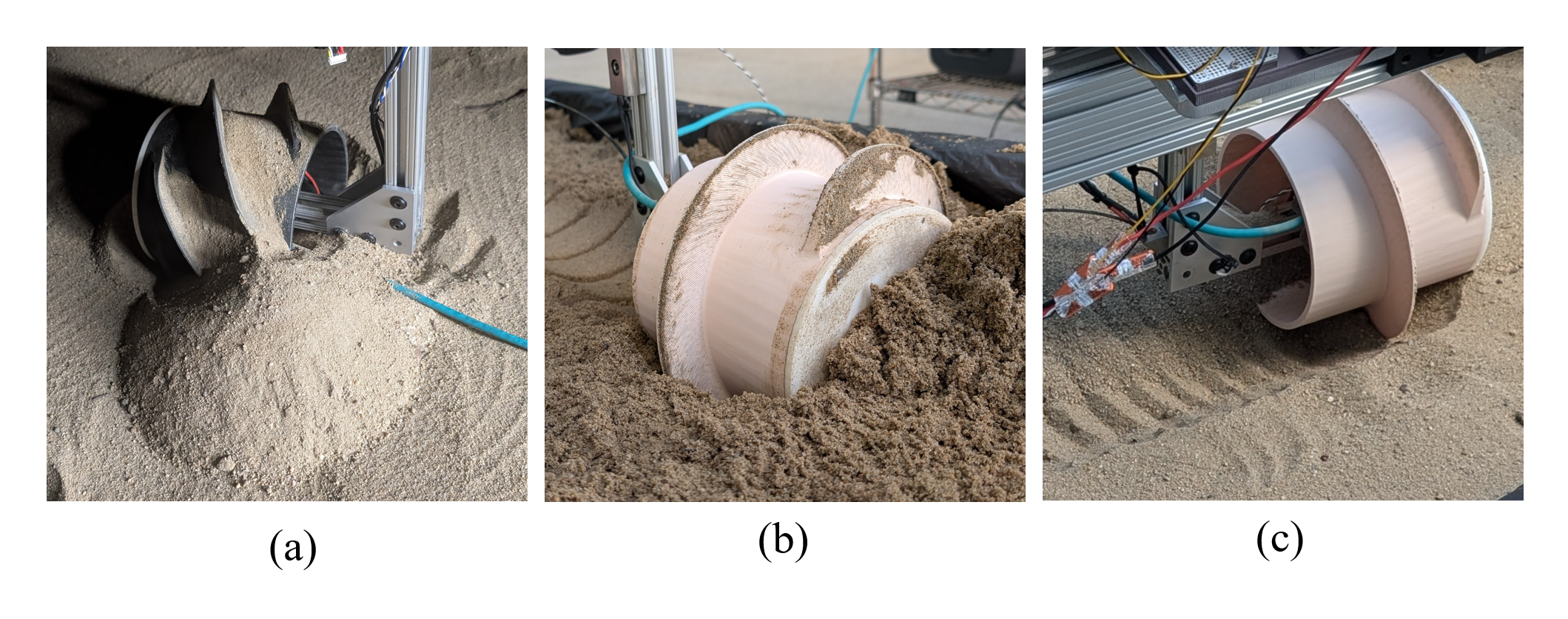}
    \caption{(a) An example of medium being pushed to the side of the screw in an "lateral displacement" effect (b) An example of tunneling where the screw is buried within the sand (c) A screw failing due to material shear failure where the screw threads go from defined to undefined at the point of failure underneath the screw}
    \label{fig: failure modes}
\end{figure}

\begin{figure*}[t]
    \vspace{2mm}
    \centering
    \includegraphics[width=0.85\linewidth]{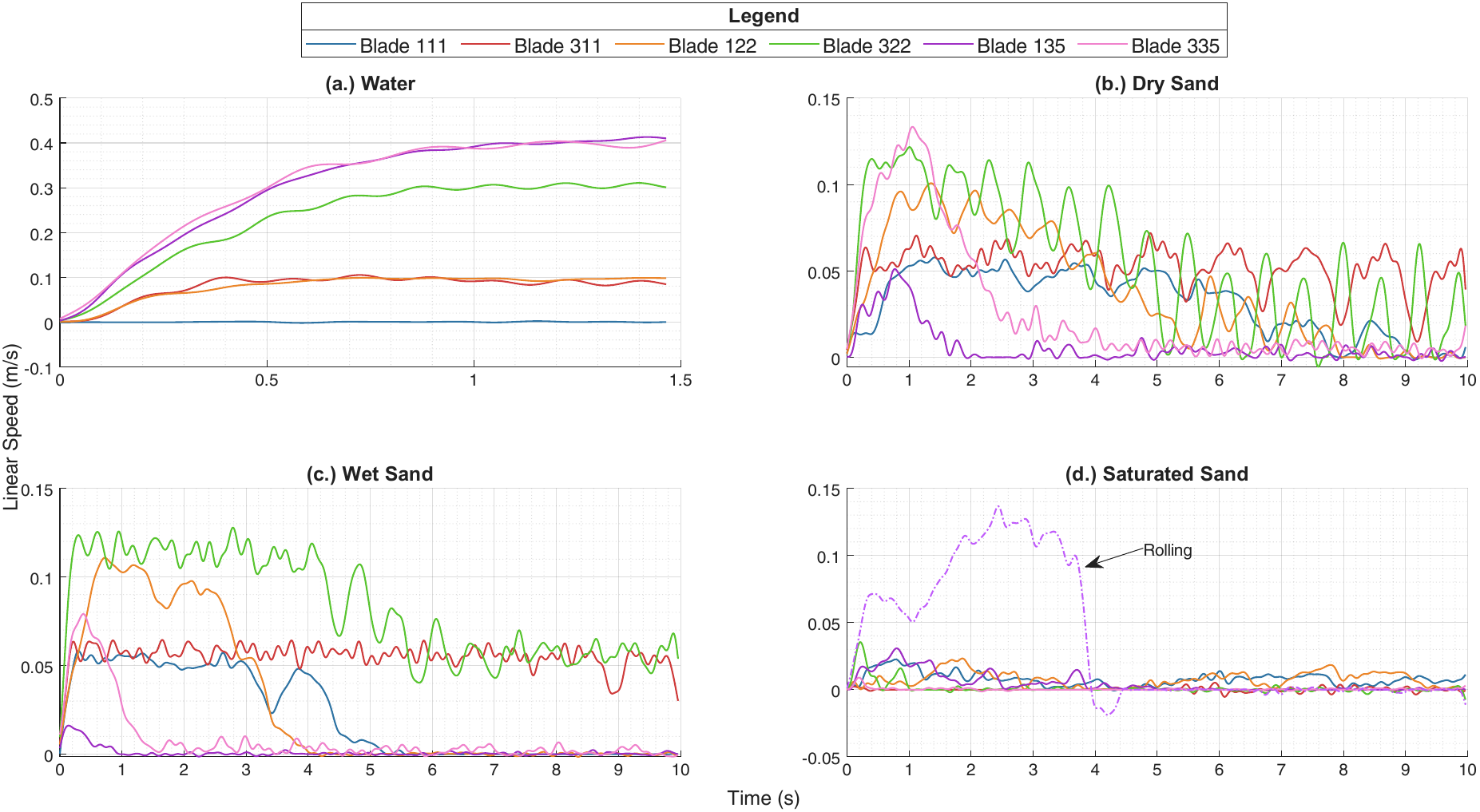}
    \caption{(a) Results of water testing with input rotational speed at a fixed 8 rad/ sec  (b) Results from dry sand testing, the rotational speed was kept at a constant 5 rad/sec (c) Results from wet sand testing with the rotational speed fixed at 5 rad/sec  (d) Results from saturated sand comparing the rolling versus screwing performance of screws at 5 rad/sec }
    \label{fig:overall_results}
\end{figure*}

\begin{figure*}[t]
    \centering
    \includegraphics[width=0.85\linewidth]{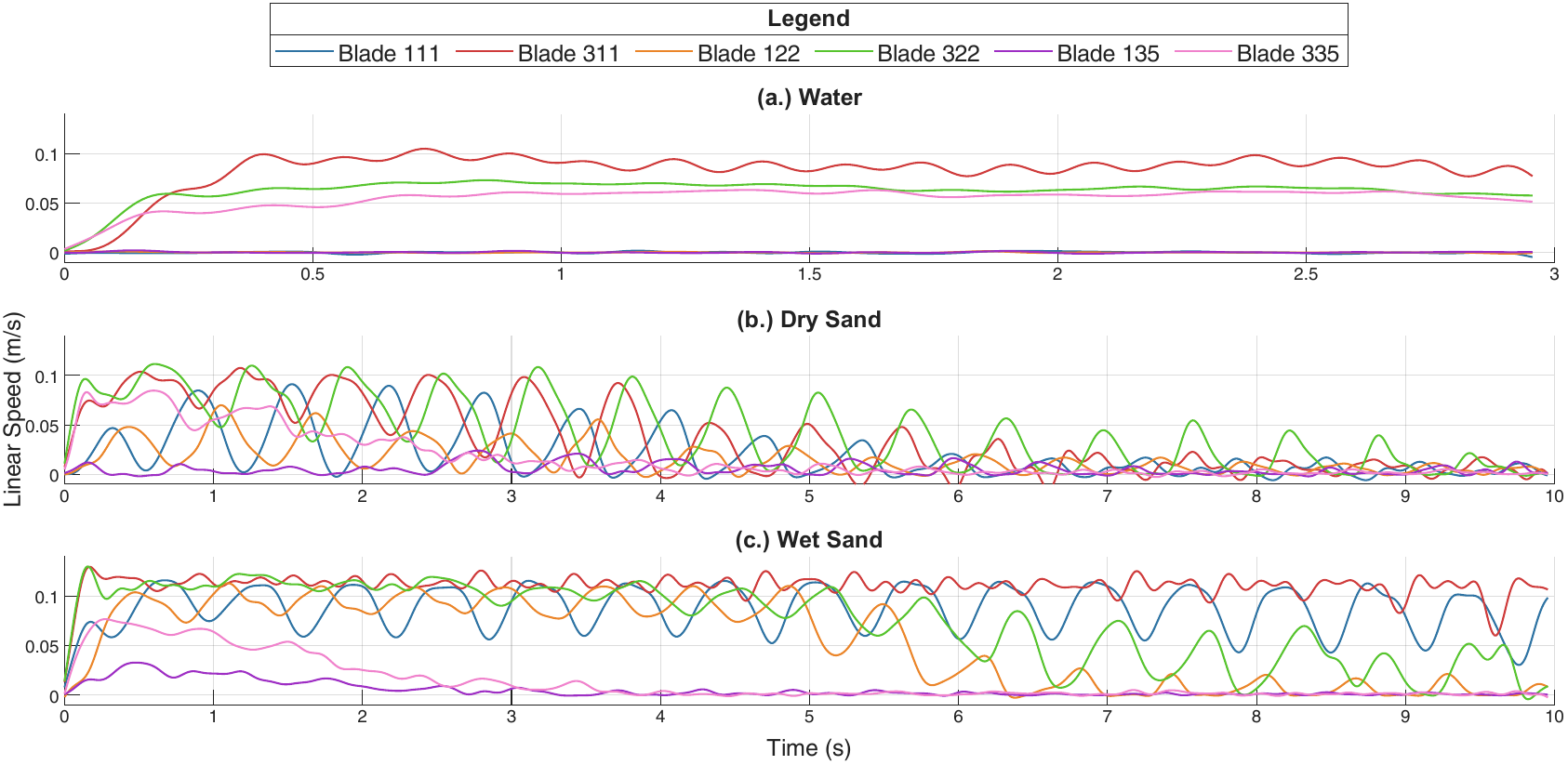}
    \caption{(a) Matched tip speed water testing, with the screw shells being run at 8, 20, and 40 radians/second, respectively, for low, medium, and high pitch heights. (b,c) Matched tip speed wet sand testing, where the low, medium, and high pitch screw shells were run at 2, 5, and 10 radians/second, respectively.}
    \label{fig:overall_results_matching}
\end{figure*}

\subsection{Experimental Set Up and Data Filtering}

Experimental data was collected on a custom-built testbed \cite{lim2023mobilityanalysisscrewbasedlocomotion} which enables pure linear motion of the screw and the collection of force and velocity measurements. New additions to enable aquatic testing, such as the newly designed section shown in Fig. \ref{fig:testsetup_temp}, as well as setup changes to enable the testing of rolling as a method of locomotion. The land-based propulsion attachments were used for the sand and wet sand testing, with the aquatic arm being used for saturated sand and water tests.
Blades were chosen so that the blade spacing ($d$) is consistent when viewed from a tangent plane sliced across the middle of the screw shell shown in Fig. \ref{fig: Derived_Design_Param}. In addition, a variable pitch blade was created to examine the effects of variable pitch across a screw shell for locomotion. Due to the difficulties in finding a representative angle for calculating tip speed, matching tip speed was done through matching revolutions over unit length. The low pitch screw shell was designed to have 0.35 revolutions, the medium 0.7 revolutions, and the high pitch with 1.75 revolutions. As such, the matching velocity was done through relative revolutions with the medium pitch screw being spun at a speed 2.5 times that of the high pitch screw, and the low pitch screw being spun at a speed 5 times that of the high pitch screw.

For the best comparison across media and different attachments, a few steps were taken to filter the data. First data from the sensors was taken in a static position to measure the noise, which was found to be ~20 Hz. The two systems also introduced some noise as the trials caused system vibrations. The regular testbed, with its shorter arm, produced higher frequency noise of ~15 Hz, while the water testbed had low frequencies in the 5 Hz and below range. Across trials, a low-pass filter with a cut-off in a range of 4 Hz - 6 Hz. 

\begin{table}[t]
\centering
\caption{Water Results}
\label{table:Water_Results}
\footnotesize 
\setlength{\tabcolsep}{4pt} 
\begin{tabular}{|l|c|l|c|}
\hline
\textbf{Shell} & \textbf{Fixed Vel. (m/s)} & \textbf{Shell} & \textbf{Matched Vel. (m/s)} \\
\hline
111 & 0.000 & 111 & 0.000\\
\hline
122 & 0.078 & 135 & 0.000 \\
\hline
311 & 0.083 & 122 & 0.000 \\
\hline
Variable & 0.204 & 335 & 0.045 \\
\hline
322 & 0.248 & 322 & 0.060 \\
\hline
335 & 0.302 & 311 & 0.083 \\
\hline
135 & 0.329 & ( - ) & ( - ) \\
\hline
\end{tabular}
\end{table}

\subsection{Aquatic Testing}

In aquatic tests, the shells that performed best (highest linear speed) typically had higher pitch angles. However, when two screw shells have the same pitch, blade height becomes the distinguishing factor in performance. In Table \ref{table:Water_Results}, of the two screw shells with high pitch angles, the low blade height configuration achieved a linear speed 8.9\% higher than the shells with high blade height. The results also showed that aspect ratio plays a role in screw effectiveness when comparing shells with similar base design parameters. Among screw shells with the same blade height, those with the lowest aspect ratios (least crowded) performed best. When comparing screw blades with the same pitch and number of starts, a similar relationship is seen in Figure \ref{fig:overall_results}, with screw shell 135 achieving higher linear speed than screw shell 335. 

The correlations between aspect ratios and performance within similar base parameters break down when pitch is too low. Screw shells 311 and 111, having the lowest pitch, do not perform well enough to provide meaningful comparative metrics. In fact, screw shell 111 does not move at all. Comparing the aspect ratio to the heat sink model, there exists an optimal geometry composed of blade spacing and blade height that allows for maximal heat dissipation or, in this case, mass transport. High blade height causes interference and larger leading vortices, decreasing screw performance in water. This leading vortex generation is often studied, as seen most recently in the context of insect and bird flight \cite{liu2024vortices}.

The matched speed data provides context for pitch angle effectiveness. Pitch angle contributes primarily to the velocity of water being pushed backwards. By matching the tip speeds of different screws, lower pitch screws (322 and 311) seen in Figure \ref{fig:overall_results_matching} prove more effective due to less water being pushed sideways, resulting in more rotational energy being converted to backward thrust.

\begin{table}[t]
\centering
\caption{Sand Results}
\label{table:Sand Results}
\footnotesize 
\setlength{\tabcolsep}{4pt} 
\begin{tabular}{|l|c|l|c|}
\hline
\textbf{Shell} & \textbf{Fixed Vel. (m/s)} & \textbf{Shell} & \textbf{Matched Vel. (m/s)} \\
\hline
135 & 0.006 & 135 & 0.005 \\
\hline
335 & 0.026 & 122 & 0.016 \\
\hline
Variable & 0.029 & 335 & 0.017 \\
\hline
111 & 0.031 & 111 & 0.02 \\
\hline
122 & 0.039 & \textit 311 & 0.036 \\
\hline
311 & 0.052 & 322 & 0.042 \\
\hline
322 & 0.056 & ( - ) & ( - ) \\
\hline
\end{tabular}
\end{table}

\subsection{Sand Testing}

For sand testing, Table \ref{table:Sand Results} shows the two best-performing screw shells had high blade heights. The third and fourth screw shells, following the top two in performance, had low blade heights. Of these four screw shells, none had high pitch angles. The screw shells with the highest pitch angle (36.2$^{\circ}$) in Table \ref{table:Sand Results} performed poorly, regardless of blade height. This contrasts with water, where high pitch angles perform well. The two screw shells that performed best in water performed worst in dry sand. Very high pitch angles in sand are unable to perform well due to the lack of reflow (material filling back under the screw). This causes the screw shell to dig and move sand to the outside of the screw rather than propel forward.

A connection to the aspect ratio can be made in sand tests. When the predominant design parameter is kept constant, in this case blade height, the aspect ratio can be used to explain differences in performance. The top two performers in sand, screw shells 322 and 311, as seen in Figure \ref{fig:overall_results}, both had high blade heights, but 322 had a higher aspect ratio and performed better. For the following two screw shells, which had low blade heights, screw shell 122 had a higher aspect ratio than screw shell 111 and performed better as well. A higher aspect ratio relates to less crowding between the blades, ensuring a cohesive packet of sand is transported as the screw shell rotates. If the aspect ratio is lower, shear forces between the blades can cause the sand to shear and fall apart, as seen in Figure \ref{fig: failure modes}. As a standard screw shell travels in the sand, it creates a groove in the sand that the threads can travel across and more easily propel the screw shell forward. As can be observed in the data, there are a number of screw designs that initially move but end up digging into the sand and halting, as seen with screws 135 and 335 in Figure \ref{fig:overall_results}. This form of shear failure causes the screw shell to lose grip and dig down into the sand instead \ref{fig: failure modes}. Similar results can be found when different types of wheels have been tested in granular media in previous works \cite{wong1967prediction}.

At matched tip speeds, similar initial performance can be observed between screws of high blade height (322, 311, 335), as seen in Figure \ref{fig:overall_results_matching}, consistent with the effects of matching tip speed. As the higher-pitched screws spun more in the sand, there was an observed sinking phenomenon, as demonstrated in Figure \ref{fig: failure modes}, as the screw pushed the granular sand to the side and the material struggled to reflow underneath to provide propulsion. The higher the pitch of the screw, the more drastic this failure mode becomes, as seen in Figure \ref{fig:overall_results_matching}, with the higher pitch flinging more sand laterally, reducing performance over time.

\subsection{Wet Sand Testing}
Similarly to dry sand, blade height emerges as the dominant design parameter for achieving optimal velocity when sufficiently low pitch angles for movement are satisfied. The two highest-performing screw shells (322 and 311) both feature high blade heights, achieving velocities of 0.083 m/s and 0.055 m/s respectively, as seen in Table \ref{table:Wet Sand Results}. This represents a substantial improvement over their dry sand counterparts, with shell 322 showing a 48\% velocity increase and shell 311 demonstrating a 6\% improvement. With water added to sand, the cohesion of the material ($c$) increases, contributing to two effects: 1) higher cohesion value ($c$) increases the material's resistance to shear failure, enabling more consistent power transmission from blade to medium and 2) allowing the helical grooves formed by the screw blades to resist collapse under subsequent cycles \ref{fig: failure modes}. The trend of higher aspect ratio to rank screw blades within the dominant design parameter of blade height can also be seen, with the higher aspect ratio screw (high blade height, medium pitch) performing better than the lower aspect ratio screw (high blade height, low pitch).

Screw shells with low and medium pitch initially demonstrated similar rotational speeds when matched for tip velocity, as seen in Figure \ref{fig:overall_results_matching}. However, over time, the velocity graph show velocity decreases as the screw tunnels due to excess sand being pushed laterally in the medium-pitched screw compared to the low pitch one. High-pitch screw shells (135 and 335) as seen in Figure \ref{fig:overall_results_matching} exhibited rapid velocity degradation over time, with their initial movement quickly transitioning to near-stall conditions. This behavior is consistent with a "lateral-displacement" phenomenon \ref{fig: failure modes}, where excessive pitch angles cause material displacement laterally rather than contributing to propulsion.

\begin{table}[t]
\vspace{2mm}
\centering
\caption{Wet Sand Results}
\label{table:Wet Sand Results}
\footnotesize 
\setlength{\tabcolsep}{4pt} 
\begin{tabular}{|l|c|l|c|}
\hline
\textbf{Shell} & \textbf{Fixed Vel. (m/s)} & \textbf{Shell} & \textbf{Matched Vel. (m/s)} \\
\hline
135 & 0.001 & 135 & 0.004\\
\hline
335 & 0.008 & 335 & 0.005\\
\hline
111 & 0.021 &  111 & 0.032\\
\hline
Variable & 0.023 & 122 & 0.036\\
\hline
122 & 0.028 & 322 & 0.063\\
\hline
311 & 0.055 &  311 & 0.066\\
\hline
322 & 0.083 & ( - ) & ( - )\\
\hline
\end{tabular}
\end{table}

\begin{table}[t]
\vspace{2mm}
\centering
\caption{Saturated Sand Results}
\label{table:Saturated Sand Results}
\footnotesize 
\setlength{\tabcolsep}{4pt} 
\begin{tabular}{|l|c|l|c|}
\hline
\textbf{Shell} & \textbf{Fixed Vel. (m/s)} & \textbf{Shell} & \textbf{Matched Vel. (m/s)} \\
\hline
311 & 0.000 & 335 & 0\\
\hline
335 & 0.000 & 135 & 0.005\\
\hline
Variable & 0.000 & 311 & 0.007\\
\hline
322 & 0.001 & 122& 0.01\\
\hline
135 & 0.005  & 322 & 0.012\\
\hline
111 & 0.008 & 111& 0.033\\
\hline
122 & 0.008 & ( - ) & ( - ) \\
\hline
135 (rolling) & 0.137 & ( - ) & ( - )\\
\hline
\end{tabular}
\end{table}

\subsection{Saturated Sand Testing}
Blades with high blade heights stalled immediately in saturated sand, as seen in Figure \ref{fig:overall_results}, demonstrating negligible forward motion. Those with lower blade heights demonstrated initial movement but stalled as well. The difference in movement can be seen in the output velocity of the low height, low pitch screw for the same input speed between Table \ref{table:Sand Results} and Table \ref{table:Saturated Sand Results} (0.031 and 0.008 m/s, respectively). With high water content, the sand becomes a non-Newtonian fluid, causing the material to behave as a solid rather than a granular substance when the blades shear into the material. Under these conditions, the screw stalls due to a torque demand exceeding available motor output. For the lower blade heights, the screws are able to get an initial motion. However, even the smallest amount of sinkage caused the screw to stall. Similar results were also seen in the matched velocity testing for Table \ref{table:Saturated Sand Results}.

Instead of shearing through the medium, more effective locomotion is demonstrated by rolling across the saturated medium, as seen in the setup for a rolling configuration in Figure \ref{fig:testsetup_temp}, where the screw can more effectively function as a wheel. Using a rolling strategy, forward velocity increases 27-fold, as seen in the comparison between the high pitch, low blade height output velocities in Table \ref{table:Saturated Sand Results}.

\subsection{Variable Parameters}
The variable blade was expected to perform similarly to the top performers, given its high blade height and low average pitch in sand. However, the variable pitch along the axis means that the screw shell must create new grooves in the sand as it travels. This has the same effect as high pitch, where the shell digs into the sand, contributing to worse performance than expected. In water, the variable pitch screw ranked as expected, given its pitch angle. With water being a liquid, the screw functions more like a propeller, with performance more closely linked to the medium being pushed back as a volume rather than screw-based propulsion.

\begin{figure}
    \centering
    \includegraphics[width=0.7\linewidth]{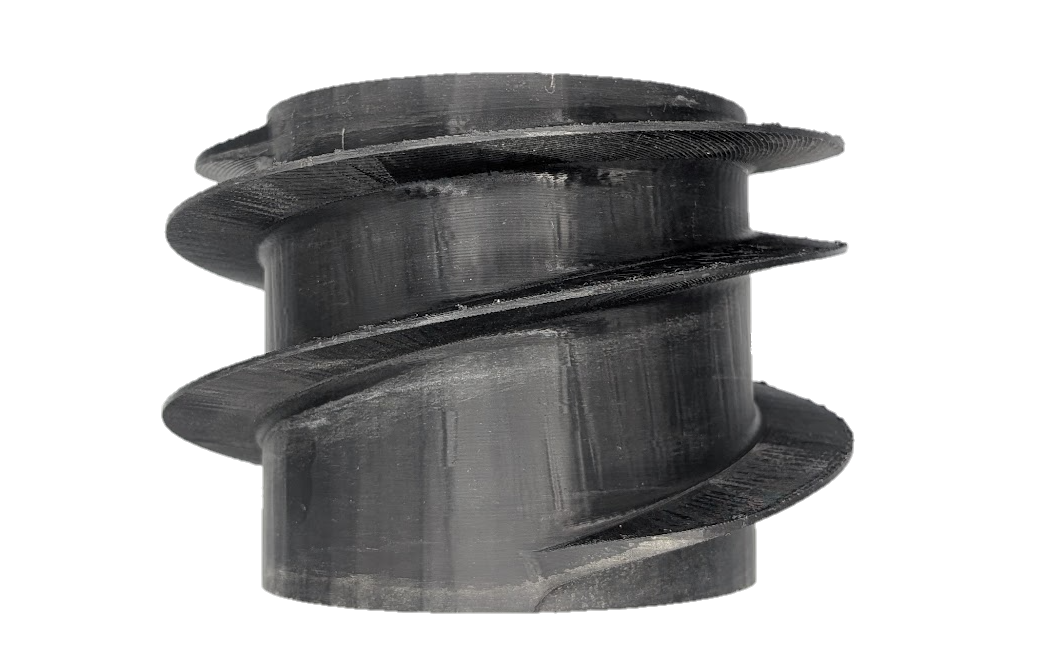}
    \caption{Example of a screw shell with pitch being changed across the length}
    \label{fig: optimizingParameter_results}
\end{figure}

\section{Discussion}

Given the complexity and number of models required to understand screw locomotion, it is best to see empirical data and build intuition on material behaviors to best choose screw parameters for any screw propelled robot.

The overall results show that in granular media, blade height provides the dominant metric for good performance, while pitch is the most effective in water. However, a good pitch in water does not result in good locomotion in sand; in fact, a high pitch performed the worst, moving very little if at all. Looking at the tip speed and taking into account the aspect ratio, a more neutral selection can be made which is capable of performing well in both, but not best in either.


\section{Conclusion}

This work presents a comprehensive experimental analysis of screw shells across distinct media, with the results of the study confirming that predominant design parameters exist for Archimedes screws in each medium (blade height in granular media, pitch in water), with aspect ratio providing a method to rank screws within dominant parameters.

It is acknowledged that these results are not a definitive model of exact screw performance in each medium. However, the experiments show trends that can be broadly applied in selecting parameters that perform well selectively in one medium and across multiple media.

In future work on screw-propelled robots for multi-media applications, there is potential for variable-speed drive trains that unlock better water and sand combined locomotion. Sand propulsion eventually proved torque-limited, while in water, faster rotation proved more effective, even generating thrust for low-pitched shells. Equally valuable is a means of sensing transitions between media, using motor current measurements to define transitions between locomotion modes to avoid stalling in media similar to saturated sand.

\section{Acknowledgments}

The Authors would like to thank Peihan Zhang, Andrew Nakoud, Florian Richter, Sara Wickenhiser, and Brian Lee.





\balance
\bibliographystyle{ieeetr}
\bibliography{refs}
\balance

\end{document}